\documentclass[runningheads]{llncs}
\let\llncssubparagraph\subparagraph
\let\subparagraph\paragraph
\usepackage[compact]{titlesec}
\let\subparagraph\llncssubparagraph
\usepackage[T1]{fontenc}
\usepackage{graphicx}
\usepackage{textcomp}
\usepackage{booktabs}
\usepackage[misc]{ifsym}
\usepackage{titlesec}
\usepackage{makecell}

\newcommand{\corr}{(\Letter)}
\usepackage{amsmath}
\usepackage{amssymb}
\usepackage{stmaryrd}
\usepackage{wrapfig}
\usepackage[table]{xcolor}
\definecolor{bluecb}{rgb}{0.45,0.31,0.49}
\definecolor{orangecb}{rgb}{0.88, 0.31, 0.24}
\definecolor{darkgray}{rgb}{0.15,0.15,0.15}
\definecolor{lightgray}{rgb}{0.94,0.94,0.94}
\definecolor{lightlightgray}{rgb}{0.97,0.97,0.97}
\definecolor{darkred}{rgb}{0.80,0.00,0.00}
\definecolor{darkgreen}{rgb}{0.00,0.70,0.00}
\definecolor{darkblue}{rgb}{0.00,0.00,0.70} 
\usepackage[square, numbers, sort&compress, comma]{natbib}
\usepackage[
    colorlinks=true,
    citecolor=black,
    linkcolor=black
    ]{hyperref}
\usepackage{stackengine}
\usepackage{tikz}                  
\usetikzlibrary{arrows,positioning,shapes}
\usetikzlibrary{shapes.multipart}
\usetikzlibrary{arrows.meta}
\usetikzlibrary{decorations.pathreplacing}
\usepackage{rotating}
\usepackage{geometry}
\usepackage{float}
\usepackage{xspace}
\usepackage{enumitem}
\usepackage{subcaption}
\usepackage{hyperref}
\newcommand{\G}{\ensuremath{\mathcal{G}}\xspace}
\newcommand{\V}{\ensuremath{\mathcal{V}}\xspace}
\newcommand{\E}{\ensuremath{\mathcal{E}}\xspace}
\newcommand{\RR}{\ensuremath{\mathbb R}\xspace}

\newcommand{\NN}{\ensuremath{\mathbb N}\xspace}
\newcommand{\EE}{\ensuremath{\mathbb E}\xspace}

\newcommand{\bX}{\ensuremath{\mathbf X}\xspace}
\newcommand{\by}{\ensuremath{\mathbf y}\xspace}
\newcommand{\bA}{\ensuremath{\mathbf A}\xspace}
\newcommand{\bW}{\ensuremath{\mathbf W}\xspace}

\newcommand{\specialcell}[2][c]{%
\begin{tabular}[#1]{@{}c@{}}#2\end{tabular}}
\newcommand{\inlinetitle}[2]{\noindent\textbf{{#1}{#2}}}

\usepackage[normalem]{ulem} 

\newcounter{marginNoteCounter}

\begin{document}

\title{Leveraging Graph Neural Networks to Forecast Electricity Consumption}

\titlerunning{Energy Forecasting with GNNs}

\author{Eloi Campagne\inst{1,3}\corr \quad
Yvenn Amara-Ouali\inst{2,3} \quad Yannig Goude\inst{2,3} \quad Argyris Kalogeratos\inst{1}}

\authorrunning{E. Campagne et al.}

\institute{Centre Borelli, Université Paris-Saclay, CNRS, Ecole Normale Supérieure Paris-Saclay – France\footnote{Contact: \email{\{name.surname\}@ens-paris-saclay.fr}}
\and
Laboratoire de Mathématiques d’Orsay (LMO), Université Paris-Saclay, CNRS, Faculté des Sciences d’Orsay – France\footnote{Contact: \email{\{name.surname\}@universite-paris-saclay.fr}}
\and
EDF R\&D, Palaiseau – France\footnote{Contact: \email{\{name.surname\}@edf.fr}}
}

\maketitle              

\begin{abstract}
Accurate electricity demand forecasting is essential for several reasons, especially as the integration of renewable energy sources and the transition to a decentralized network paradigm introduce greater complexity and uncertainty. The proposed methodology leverages graph-based representations to effectively capture the spatial distribution and relational intricacies inherent in this decentralized network structure. This research work offers a novel approach that extends beyond the conventional Generalized Additive Model framework by considering models like Graph Convolutional Networks or Graph SAGE. These graph-based models enable the incorporation of various levels of interconnectedness and information sharing among nodes, where each node corresponds to the combined load (i.e. consumption) of a subset of consumers (e.g. the regions of a country). More specifically, we introduce a range of methods for inferring graphs tailored to consumption forecasting, along with a framework for evaluating the developed models in terms of both performance and explainability. We conduct experiments on electricity 
 forecasting, in both a synthetic and a real framework considering the French mainland regions
, and the performance and merits of our approach are discussed.

\end{abstract}

\section{Introduction}

The effective operation of the electrical system relies on maintaining a balance between electricity supply and demand. Since electricity cannot be efficiently stored, its production needs to be constantly adjusted to match consumption. Providing accurate forecasts for short-term electricity demand is therefore crucial for all participants in the energy market. The shift toward a decentralized electricity network introduces new uncertainties, which pose additional challenges for demand forecasting. In addition, the increasing contribution of renewable energy sources, like solar and wind power, brings fluctuations and intermittency to the electricity market. These fluctuations and intermittency occur at various spatial scales due to the presence of wind farms and photovoltaic power plants. The crisis brought by Covid-19, along with the current economic downturn, add further complexity to forecasting due to the non-stationarity in consumption patterns \citep{alasali2021impact}. The availability of new geolocated data and individual electricity consumption data can be exploited by models that are able to take advantage of additional information and help in minimizing forecast uncertainty \citep{adaptive_methods, devilmarest2021statespace}. Furthermore, recent advancements in adaptive forecasting algorithms have demonstrated improvements in forecasting quality, particularly for aggregate load forecasting \citep{bregere2022online, antoniadis2022hierarchical}. 

\smallskip
\inlinetitle{Related Work}{.}~%
Forecasting models are essential for estimating quantities such as electricity demand and renewable energy production. Accurate predictions of these quantities are crucial for decisions related to energy market positioning and grid optimization. Probabilistic forecasting methods have been extensively studied in the context of electricity demand. Generalized Additive Models (GAMs) have shown strong performance in conditional mean forecasting, as demonstrated by \cite{fan2010forecast} and \cite{pierrot2011short}. Building on this, \cite{fasiolo2021fast} and \cite{Gilbert2023} introduced quantile GAMs and a variant of Generalized Additive Models for Location Scale and Shape (GAMLSS) to model the full distribution of demand.
While these models are effective for stationary data, electricity consumption patterns often change over time due to factors such as the Covid-19 pandemic, which altered electricity usage habits \cite{jiang2021impacts}, and the recent surge in electricity prices in Europe, which led to reduced consumption \cite{doumeche2023human}. Traditional offline quantile regression methods are inadequate for capturing such dynamic changes. To address this, \cite{de_vilmarest_adaptive_2023} proposed adaptive probabilistic forecasting methods, incorporating the Kalman filter with GAMs and Online Gradient Descent, to better handle evolving electricity data.
Therefore, modeling national net demand can be achieved with a wide variety of methods working directly on this quantity. However, a decentralized electricity grid calls for decentralized methods to forecast net demand with the shift toward a smart grid paradigm \cite{williams2020electricity}.

Deep learning methods have shown great flexibility and adaptability to a wide range of problems applied to the energy sector \cite{9395437}. To outperform standard methods, they usually require a large amount of data along with intricate relationships between the input features and the target. These conditions are met particularly in decentralized grids, where multiple forecasting problems can be factored with the adequate neural network architecture. Recurrent neural networks (RNNs), particularly Long Short-Term Memory (LSTM) networks, have been widely used in electricity forecasting due to their 
capacity to model temporal dependencies. \cite{shi2017deep} proposed a Deep LSTM model for short-term household load forecasting, which showed superior performance over traditional statistical methods such as ARIMA. Similarly, \cite{marino2016building} developed a sequence-to-sequence LSTM model that captures the sequential nature of load data effectively.
Graph Neural Networks (GNNs) have been increasingly applied to time-series forecasting tasks, demonstrating their effectiveness in capturing complex dependencies within temporal data. \cite{guo2019attention} introduced an attention-based spatio-temporal GNN for traffic forecasting, leveraging the attention mechanism to dynamically capture the importance of different spatial and temporal features, which significantly improved forecasting accuracy. Similarly, \cite{chen2020dynamic} developed a dynamic spatio-temporal GNN for traffic flow prediction, where the model adaptively extracted features through a dynamically constructed graph. In the context of load forecasting, \cite{jiang2023buaa_bigscity} proposed a hybrid GNN framework to capture the intricate relationships between different nodes in a power grid, resulting in superior performance over traditional methods for the Baidu KDD Cup 2022 Spatial Dynamic Wind Power Forecasting Challenge. Another notable contribution by \cite{JIANG2024123435} involves a comprehensive GNN-based approach for energy load forecasting, integrating various external factors to enhance prediction robustness. These advancements underscore the potential of GNNs to model and forecast complex time-series data effectively, establishing them as a valuable asset in load forecasting. The primary challenge lies in designing problem-specific graphs, which requires leveraging the experience and expertise acquired in the field. 

Combining traditional statistical models with modern machine learning techniques has also shown promising results. Neural Additive Models (NAMs) \cite{agarwal2021neuraladditivemodelsinterpretable} combine the interpretability of GAMs with the flexibility of neural networks, providing a powerful tool for regression tasks in various domains, including energy forecasting. Furthermore, Neural Additive Models for Location Scale and Shape (NAMLSS) \cite{thielmann2024neuraladditivemodelslocation} extend GAMLSS by incorporating neural network components, allowing for more complex and adaptive distributional modeling.

\smallskip
\inlinetitle{Contribution}{.}~%
In this paper, we present how GNNs can be applied to electricity load forecasting. In particular, we show different techniques that allow to effectively build representative graphs in the electricity forecasting context, and combine them to get the most out of those techniques. In order to better understand the behavior and then benchmark different GNN models, we introduce a new synthetic dataset with added correlations between the nodes that we seek to retrieve in output. To complete this study, we apply an explainability algorithm developed in \cite{ying2019gnnexplainer} which extracts important subgraphs by maximizing a mutual information criterion.

\smallskip
\inlinetitle{Paper organization}{.}~%
Section 2 reviews the main tools that will be employed later in this study; in Section 3 we develop our forecasting approach for energy consumption; in Section 4, we show comparative results about its performance when used in practice. Finally, we discuss our findings in Section 5, and we give our conclusive remarks in Section 6.
%
%

\section{Preliminaries and Background}

In this section, we introduce the basics of GNNs, additive models and expert aggregation.

\subsection{Graph Neural Networks}
In graph theory, objects are represented by nodes, and the relationships between them are represented by edges. Formally, a graph $\G = (\V, \E)$ is a couple where \V is a set of nodes and \E a set of edges, i.e. $\E = \{e_{ij} = v_iv_j ~|~ v_i, v_j \in \V \}$.  A graph can be represented either using a binary-weighted adjacency matrix $\bA = (\bA_{ij}) \in \RR^{N\times N}$ such that $\bA_{ij} = 1$ if and only if $e_{ij} \in \E$, or a more flexible real-valued weight matrix $\bW$. We denote the neighborhood of node $v$ by the set $\mathcal{N}_v=\{u \in \V,\, \bA_{uv} = 1\}$, or equivalently $\{u \in \V,~ \bW_{uv} > 0\}$. Note that we consider symmetric graphs, therefore $\bA = \bA^\top$ and $\bW = \bW^\top$. In the context of electricity load forecasting in France, regions and the unknown links between them can respectively be seen as nodes and edges. Both regression and classification tasks can be performed on graphs at different levels: node-level, edge-level, graph-level.
%

\begin{itemize}[topsep=0.4em, itemsep=0.em]
    \item The \textbf{node-level} focuses on individual nodes within a graph. It involves analyzing the properties or attributes of each node. For example, in the context of electricity forecasting, a node-level task could be to predict the consumption for each region. 
    \item The \textbf{edge-level} 
    pertains to the analysis of the edges or connections between nodes in a graph. It involves examining the relationships, weights, or properties associated with each edge. For example, in the context of electricity forecasting, an edge-level task could be to quantify the relationships between the regions.
    \item The \textbf{graph-level} refers to the analysis of the entire graph structure as a whole. It involves examining global properties, overall connectivity, or emergent behaviors of the graph. For example, in the context of electricity forecasing, a graph-level task could be to predict the national consumption.
\end{itemize}

\begin{figure}[t!]
    \centering
    \scalebox{.8}{\begin{tikzpicture}[node distance=2cm]
        \begin{scope}[xshift=-1.5cm]
            \draw[fill=gray!20, rounded corners] (-0.7,-5.7) rectangle (6,1);

            \node[draw, circle, fill=gray!10] (Node_n) {$V_n$};
            \node[draw, circle, fill=gray!10, below of=Node_n] (Edge_n) {$E_n$};
            \node[draw, circle, fill=gray!10, below of=Edge_n] (Global_n) {$U_n$};
              
            \node[draw, circle, fill=gray!50, right=4cm of Node_n] (Node_np1) {$V_{n+1}$};
            \node[draw, circle, fill=gray!50, below of=Node_np1] (Edge_np1) {$E_{n+1}$};
            \node[draw, circle, fill=gray!50, below of=Edge_np1] (Global_np1) {$U_{n+1}$};
              
            \draw[->, dashed, darkblue] (Node_n) -- node[midway, above] {$\phi^v$} (Node_np1);
            \draw[->, dashed, darkblue] (Edge_n) -- node[midway, above] {$\phi^e$} (Edge_np1);
            \draw[->, dashed, darkblue] (Global_n) -- node[midway, above] {$\phi^u$} (Global_np1);

            \draw[- Circle, dotted, orange] (0.5, 0) to [bend right=50] node[midway, left] {$\rho_{V_n\to E_n}$} (2, -2.);
            \draw[-, dotted] (0.5, -4) to [bend left=50](2, -2.);
            \draw[-, dotted] (0.5, -4) to [bend right=50] (4, 0.);
            \draw[- Circle, dotted, red] (2, -2) to [bend left=50] node[midway, right] {$\rho_{E_n\to V_n}$} (4, 0.);
            \draw[- Circle, dotted, darkgreen] (2, -2) to [bend right=50] node[midway, right] {$\rho_{E_n\to U_n}$} (4, -4);
            \draw[-, dotted] (0.5, 0) to [bend left=50] (4, -4.);

            \node[below] at (2.5, -5) {\textit{Message passing layer}};
        \end{scope}
    
        \begin{scope}[xshift=-6cm, yshift=-3cm]
            \draw[dashed, rounded corners] (-0.7,-1) rectangle (2.7,3);
            
            \node[circle, draw, fill=blue!20] (Left_A) at (0,0) {D};
            \node[circle, draw, fill=red!20] (Left_B) at (2,0) {C};
            \node[circle, draw, fill=green!20] (Left_C) at (2,2) {B};
            \node[circle, draw, fill=orange!20] (Left_D) at (0,2) {A};
              
            \draw[red] (Left_A) -- (Left_B);
            \draw[green] (Left_B) -- (Left_C);
            \draw[orange] (Left_C) -- (Left_D);
            \draw[blue] (Left_D) -- (Left_A);
            
            \draw[-] (2.7,-1) -- (3.8,-2.6);
            \draw[-] (2.7,3) -- (3.8,3.93);
        \end{scope}
        
        \begin{scope}[xshift=6cm, yshift=-3cm]
            \draw[dashed, rounded corners, line width=2pt] (-0.7,-1) rectangle (2.7,3);
            
            \node[circle, line width=2pt, draw, fill=blue!60!red] (Right_A) at (0,0) {D};
            \node[circle, line width=2pt, draw, fill=red!60!green] (Right_B) at (2,0) {C};
            \node[circle, line width=2pt, draw, fill=green!60!orange] (Right_C) at (2,2) {B};
            \node[circle, line width=2pt, draw, fill=orange!60!blue] (Right_D) at (0,2) {A};
              
            \draw[red, line width=2pt] (Right_A) -- (Right_B);
            \draw[green, line width=2pt] (Right_B) -- (Right_C);
            \draw[orange, line width=2pt] (Right_C) -- (Right_D);
            \draw[blue, line width=2pt] (Right_D) -- (Right_A);

            \draw[-] (-1.5,-2.6) -- (-0.7,-1);
            \draw[-] (-1.5,3.93) -- (-0.7,3);
        \end{scope}
    \end{tikzpicture}}
    \caption{Example of a message passing layer in a GNN. $V_n$, $E_n$ and $U_n$ respectively refer to node, edge, and global level at stage $n$. $\phi$ are update functions and $\rho$ are propagation functions.}
    \label{fig: message_passing}
\end{figure}
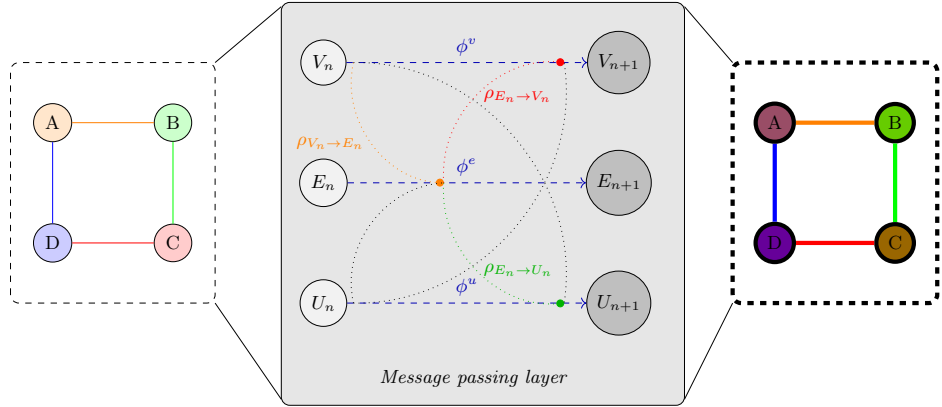

In the context of GNNs, message passing refers to the process of exchanging information between nodes, edges, and the global level of a graph, see Figure \ref{fig: message_passing}. Message passing is a fundamental operation in GNNs that enables nodes to gather and aggregate information from their neighbors, incorporate it into their own representations, and propagate it throughout the graph.
Hence, a GNN corresponds to a set of layers that use the message-passing mechanism. Node representations are therefore updated as the graph is iterated through (in other words, at each layer traversed, representations are updated).
A message passing layer therefore enables a node to update its embedding by taking into account information from its neighbors: thus, by propagation, $d$ message passing layers enable a node to take into account information from its $d^{th}$-order neighbors. Using the notations of Figure \ref{fig: message_passing}, the message passing mechanism for each level can be written as follows:

\begin{itemize}[topsep=0.4em, itemsep=0.em]
    \item $V_{n+1} = \phi^v\left(V_n ~;~ \rho_{E_n \to V_n}, \rho_{U_n \to V_n}\right)$
    \item $E_{n+1} = \phi^e\left(E_n ~;~ \rho_{V_n \to E_n}, \rho_{U_n \to E_n}\right)$
    \item $U_{n+1} = \phi^u\left(U_n ~;~ \rho_{V_n \to U_n}, \rho_{E_n \to U_n}\right)$
\end{itemize}
In a standard neural network (e.g. a feed-forward neural network), the update process is typically a local operation. In contrast, message passing in GNNs is a more dynamic and relational process allowing nodes to capture both local and global effects and dependencies over the graph structure. This relational approach enables GNNs to capture complex graph patterns and dependencies that cannot be easily captured by standard neural networks, which makes them suitable for tasks involving graph-structured data. The message passing mechanism can also be seen as a special case of graph convolution. Convolutional operations, originally developed for regular grid-structured data such as images, enable the extraction of meaningful features from the local neighborhood of each element \citep{lecun1995convolutional}. 
By extending convolution to graphs, we can capture and analyze the structural patterns and relationships present in graph data; see \citep{daigavane2021understanding} for a visual representation. Graph convolution extracts local features by aggregating information from neighboring nodes to compute features for a node, hence allowing to capture the local connectivity and dependencies between nodes. It also helps to leverage the inherent structure and connectivity of the graph data, uncover hidden patterns (e.g. the relationship between the consumptions of two different regions), and make informed predictions (e.g. the consumption of a given region) based on the relationships between nodes. 

Let $\mathcal G = (\V, \E)$ be a graph, and for each node $v \in \V$ we associate a feature vector $\bX_v$ corresponding to the explanatory variables of the associated node. Each node $v \in \V$ has an associated label $\by_v$ and we want to learn a representation such that, for a GNN $\Phi_\theta$, we have $\Phi_\theta(\G, \bX_v) = \hat\by_v$, where $\theta$ is a vector of parameters. To do this, we iteratively update the representations of a node by aggregating the representations of its neighbors in the graph. The representation of a node $v$ at layer $\ell$, denoted $h_v^{(\ell)}$ can be expressed as: 
\begin{equation}
    \begin{aligned}
        h_v^{(\ell)} &= \mathsf{UPDATE}^{(\ell)}\left(h_v^{(\ell-1)};~\mathsf{AGGREGATE}^{(\ell)}\left(h_v^{(\ell-1)} ;\left\{h_u^{(\ell-1)} ~\big|~ u \in \mathcal N_v\right\}\right)\right), \hspace{.5cm} \\
        h_v^{(0)} &= \bX_v,
    \end{aligned}
    \label{eq: representation_equation}
\end{equation}
where \textsf{UPDATE} usually involves combining the prior representations with the current one and a linear mapping. \textsf{AGGREGATE} is usually a combination of a pooling function (e.g. $\max, ~ \mathrm{sum}$,...) with an activation function (e.g. ReLU, $\tanh$,...).



\smallskip
 \inlinetitle{Classical GNN Models}{.}~ Two of the most prominent models are the Graph Convolutional Network (GCN) and the Graph SAGE (SAmple and aggreGatE). GCNs, introduced by \cite{kipf2017semisupervised}, generalize the convolution operation to graph data by aggregating feature information from a node’s local neighborhood. In contrast, Graph SAGE, developed by \cite{hamilton2018inductive}, focuses on the inductive capability of GNNs by learning a function to generate embeddings for nodes based on their local neighborhood. These models have paved the way for numerous advancements in the field of graph representation learning. In the following update rules, we will use the notations $\bW^{(\ell)}$ and $\mathbf b$ respectively referring to a learnable weight matrix at layer $\ell$ and to a bias term, $\sigma$ being an activation function (in the experiments, we will use ReLU). The update rule of GCN and SAGE for node $v_i$ is given by $h_i^{(\ell+1)} = \sigma\left(\sum_{j\in\mathcal N_i}\frac{1}{c_{ij}}\bW^{(\ell)}h_j^{(\ell)} + \mathbf b\right)$, where $c_{ij} = \sqrt{|\mathcal N_{v_i}|}\sqrt{|\mathcal N_{v_j}|}$ for a GCN, and $c_{ij} = |\mathcal N_{v_i}|$ for a SAGE. In the original paper, they introduce a max-pooling aggregation method for which the update rule is given by $h_i^{(\ell+1)} = \sigma\left(\bW^{(\ell)}\left[h_i^{(\ell)}\big{|}\big{|}\max\left\{\sigma\left(\bW_{\text{pool}}h_j^{(\ell)} + \mathbf b\right),~\forall v_j\in\mathcal N_{v_i}\right\}\right]\right)$, where $\bW_{\text{pool}}$ is a learnable pooling weight matrix, which is what we will use in the experiments, and $[\cdot||\cdot]$ is a concatenation operator.

\subsection{Generalized Additive Models}
GAMs is a class of semi-parametric regression models that was developed in \citep{gam1, wood2017generalized} and are now widely used in electricity consumption forecasting \cite{gaillard2016additive}. 
Indeed, GAMs are interesting in practice, since their additive aspect makes them highly explainable, but this also means that the choice of variables must be meticulous.
Consider a prediction model aiming to predict for each time $t$ a variable of interest $y_t$ using $(x_j)_{j=1\dots d}$ explanatory variables such that $y_t = \beta_0 + \sum_{j=1}^d f_j(x_{t,j}) + \varepsilon_t$, where $\beta_0$ is the intercept and $(\varepsilon_t)$ is independent and identically distributed random variable. Here we consider that each non-linear effect $f_j$ is decomposed on a spline basis $(B_{j,k})$ with coefficients $\boldsymbol{\beta}_{j}$ where $m_j$ is the chosen spline basis dimension, such that $\textstyle f_j(x)= \sum_{k=1}^{m_j} \beta_{j,k}B_{j,k}(x)$. These coefficients are then estimated by minimizing the ridge-regression criterion ensuring the smoothness of the functions $f_j$ by controlling the second derivatives \citep{wood2016smoothing}.

\subsection{Aggregation of Experts}

Several models have been developed in the literature, each with its own distinctive features, which may also complement each other. Expert aggregation is an ensemble technique that allows to benefit from the advantages of each model: we can combine them using robust online aggregation of experts, as developed in \citep{online-agg}. For each instant in the prediction, a weight is assigned to each expert according to its previous forecasts: the better the past forecasts, the greater the weight at time $t$. Let $x_{j,t}$ be the $j^{th}$ expert at time $t$ and $p_{j,t}$ its corresponding weight, then the expression of the predicted load at time $t$ is given by $\textstyle \widehat y_t = \sum_{j=1}^K p_{j,t}x_{j,t}$, where $K$ is the number of experts in the mixture. One way to compute the weights is to use polynomially weighted averages with multiple learning rate (ML-Poly), an algorithm developed in \citep{ml-poly}. A key advantage lies in the upper bound of the algorithm's average error:%
\begin{equation}
    \stackMath
    \stackunder{\textbf{Average error}}{\textbf{of the algorithm}} \lesssim
    \stackMath
    \stackunder{\textbf{Average error of the}}{\textbf{best combination of experts}} + \sqrt{\dfrac{\textbf{Number of experts}}{\textbf{Number of days}}}.
\end{equation}
The 
average performance will converge over time to match the performance of the best experts.

\section{Developing Robust Models with GNNs}

This section presents our methodology for building robust GNNs for load forecasting: from the inference of suitable graphs, to the explanation of the forecasts obtained by the models.

\subsection{Inferring Graphs from Data}

\begin{figure}[t!]
    \centering
    \includegraphics[width=.8\textwidth]{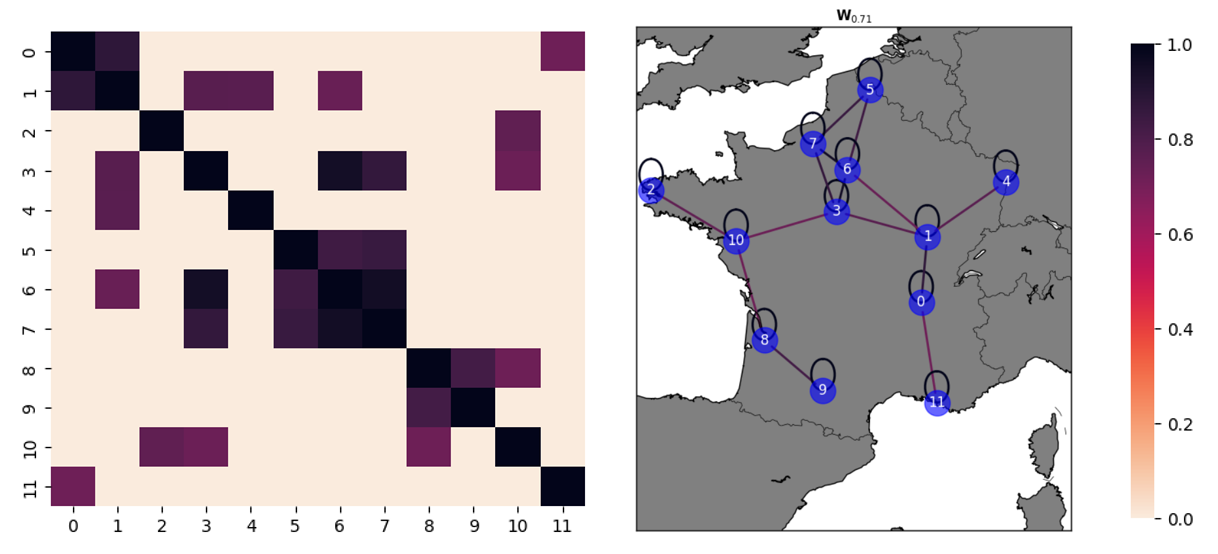}
    \caption{Graph corresponding to $\boldsymbol{W}_{\lambda}$ with $\lambda = 0.71$ and kernel bandwidth $\sigma=478.3$.}
    \label{fig: exponential-weight-matrix}
\end{figure}

\inlinetitle{Geographical Data}{.}~%
To capture the relationships between the nodes of a graph, a first approach is to study the similarity matrix of the geographical positions of these nodes. We can therefore define the weight matrix as follows:
\begin{align}
    \boldsymbol{W}_\lambda = (\boldsymbol{W}_{i,j})_{1\le i,j \le 12} = \left\{
    \begin{array}{ll}
        \exp\left\{-\dfrac{\mathbf{dist}(i,j)^2}{\sigma^2}\right\} & \mbox{if}~ \exp\left\{-\dfrac{\mathbf{dist}(i,j)^2}{\sigma^2}\right\} \ge \lambda, \\
        0 & \mbox{otherwise.}
    \end{array}
\right.
\end{align}
where $\lambda \in (0,1)$ is a threshold that controls the connectivity of the graph, $\mathbf{dist}(i,j)$ is the geodesic distance between nodes $i$ and $j$, and $\sigma$ is a bandwidth term. In practice, $\lambda$ is picked such that the graph induced by $\boldsymbol{W}_\lambda$ is minimally connected, and $\sigma$ is the median of all the distances. Other heuristics can be used to calculate the bandwidth of the Gaussian kernel, but in practice this one gives good results and is robust to outliers \cite{long2015learningtransferablefeaturesdeep, garreau2018largesampleanalysismedian}.
Figure \ref{fig: exponential-weight-matrix} shows the spatial matrix corresponding to $\lambda = 0.71$ and $\sigma = 478.3$. Note that this approach does not take geographic distances into account: we are using a simple approach here, but we could consider reducing the weights of regions separated by physical obstacles (sea, mountains, etc.). A slightly different approach might be to merge the spatial graph as presented above, with a graph encoding the different geographical clusters of the graph ($1$ if nodes belong to the same cluster, $0$ otherwise). We could, for example, consider coastal, mountain, urban and rural areas.


\begin{figure}[t]
    \centering
    \includegraphics
		[width=\textwidth]{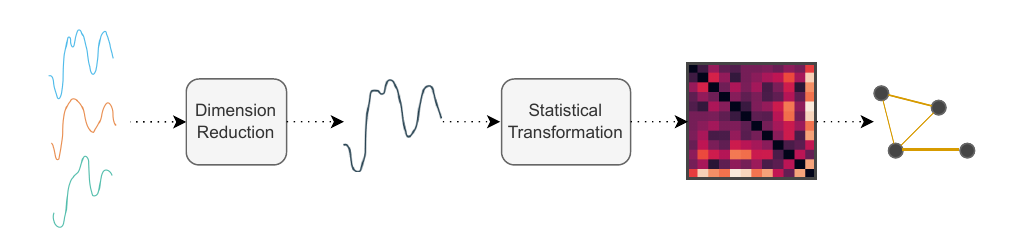}
    \caption{Inferring a graph from data using a dimension reduction algorithm and a statistical transformation.}
    \label{fig: graphdata}
\end{figure}

\smallskip
\inlinetitle{Electricity and Weather Data}{.}~%
Another way of capturing relationships between nodes is to look at spatio-temporal nodal characteristics, such as temperature, cloud cover or wind signals, which are used in practice in load forecasting. 
In practice, to build the graphs, we group all the scaled electricity and weather signals by region and we construct $n$ matrices of dimension $(d, T)$, where $n$ is the number of nodes in the graph, $d$ is the dimension of the feature space, and $T$ the number of instants in the time-series, which we project using a singular value decomposition (SVD) into a space of dimension $(1, T)$. In this way, we end up with a global matrix of dimension $(n, T)$ on which we can apply various algorithms, see Figure \ref{fig: graphdata}.
In the study, we computed the DTW distance with FastDTW \cite{salvador2004fastdtw}, an approximate Dynamic Time Warping algorithm that offers near-optimal alignments with linear time and memory complexity, between the reduced regional signals. We also tried a slightly different approach that involves estimating regional temperature effects on load (using splines, for example) and then calculating a matrix of correlated distances (such as the $L^2$-distance) between the different splines. We also implemented the GL3SR algorithm \cite{humbert2021learning}, based on two key assumptions about the signal: first, that the signal is \emph{smooth} with respect to relative to the graph structure (i.e. adjacent nodes have similar values, which is the case for weather and load signals across the mainland France), and second, that the signal is bandlimited and therefore has a sparse representation in the spectral domain.


\subsection{Explainability}

GAMs have been used for electricity forecasting because they have the advantage of being explainable. With GNNs we want to know  the hidden links between the different regions. This subject has been studied in \cite{ying2019gnnexplainer} with subgraphs, where the $\mathsf{GNNExplainer}$ algorithm was developed, which aims at explaining both the graph structure and the features. The idea is to differentiate the subgraphs that are useful in the prediction from the ones that are not. Thus, let us write $\G = \G_S + \Delta \G$ where $\G_S$ is the explaining subgraph, and $\Delta \G$ the subgraph with irrelevant edges. Then, we want to maximize the mutual information ($\mathsf{MI}$) between the predictions $\boldsymbol{Y}_\G$ and the subgraph $\G_S$ that is given by solving the optimization problem $\max_{\G_S}~\mathsf{MI}(\boldsymbol{Y}_\G,\G_S) = H(\boldsymbol{Y}_\G) - H(\boldsymbol{Y}_\G~|~\G=\G_S,~ \bX=\bX_S)$, where $H$ is the entropy function and $\bX_S$ is a subset of explaining features. In the previous equation, $H(\boldsymbol{Y}_\G)$ is constant therefore the optimization problem is equivalent to finding the minimum of the conditional entropy $H(\boldsymbol{Y}_\G~|~\G=\G_S,~ \bX=\bX_S)$, which amount to minimizing the uncertainty of the prediction $\boldsymbol{Y}_\G$. Supposing that $\mathcal G_S$ is a random graph variable following $f_{G}$, the algorithm seeks to minimize $H(\boldsymbol{Y}_\G ~|~ \G = \EE_{f_G}[\G_S],~ \bX = \bX_S)$, where $\EE_{f_G}[\G_S]$ is computed with a mean-field approximation. In other words, the objective is to pinpoint a compact subgraph along with a limited set of node attributes that greatly enhance a GNN's prediction certainty. 

\section{Experiments}

In this section, we outline the experimental procedure used to assess GNNs, covering everything from dataset preparation to result analysis, including model parameterization.

\subsection{Datasets}
We present the datasets used in our experiments: in particular we explain the generation of synthetic data and describe the real data. For both synthetic and real datasets, we consider the $12$ regions of French mainland -- excluding Corsica.


\smallskip
\inlinetitle{Synthetic Dataset}{.}~%
To evaluate the GNNs performance, we synthetically generate data. In this way, we can manually add spatial and temporal correlations between different graph nodes and obtain a lower bound on the model performance. Let us denote by $T_j^\textbf{gen}(t)$ and $L_j^\textbf{gen}(t)$ (respectively $T_j^\textbf{obs}(t)$ and $L_j^\textbf{obs}(t)$) the generated temperature and load (respectively the observerd temperature and load) at time $t$ and node $j$. We have $T_j^\textbf{gen}(t) = at + b_j(\cos{\omega_1 t} + \cos{\omega_2 t})$, where $a$ is a trend coefficient, $\omega_1$ and $\omega_2$ being respectively a daily and a yearly pulse, $b_j$ a random term such that $\mathbf b = (b_j)_{1\le j \le 12} \sim \mathcal N(\hat{\boldsymbol{\mu}}, \hat{\boldsymbol C})$ with $\hat{\boldsymbol{\mu}}$ and $\hat{\boldsymbol C}$ being respectively the empirical mean and covariance of the observed temperatures. Temperatures are then rescaled to ensure that they lie between the observed minima and maxima of each region. Then, we fit a cubic spline basis \cite{wood2002gams} such that $\tilde{f}_j\in 
{\arg\min}_{f_j}\left(f_j(T_j^\textbf{obs}) - L_j^\textbf{obs}\right)^2$ with $f_j \in \mathbf{span}(s_{j,1},\dots,s_{j,k})$ and $k$ being the number of knots. It follows that $L_j^\textbf{gen}(t) = \tilde f_j(T_j^\textbf{gen})(t)  + \varepsilon_j(t)$, where $\boldsymbol\varepsilon = (\varepsilon_j)_{1\le j\le12} \sim \mathcal N(\mathbf 0, \mathbf \Sigma)$. We simulate two datasets: one with $\mathbf{\Sigma} = \boldsymbol{\rho}(\mathbf W_\lambda)$ associated to pairwise influence between the regions, and one with $\mathbf \Sigma = \mathbf{I}$ associated to independent regions.


\begin{figure}[h!]
    \centering
    \begin{subfigure}[b]{.4\textwidth}
        \centering
        \includegraphics[width=\textwidth]{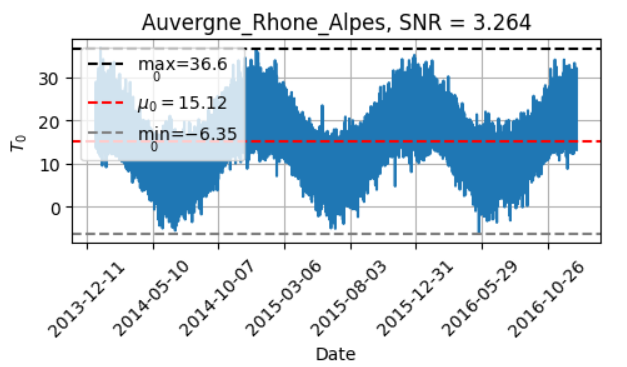}
        \caption{Temperature generated in Auvergne-Rhône-Alpes.}
    \end{subfigure}
    \hspace{.5cm}
    \begin{subfigure}[b]{.4\textwidth}
        \centering
        \includegraphics[width=\textwidth]{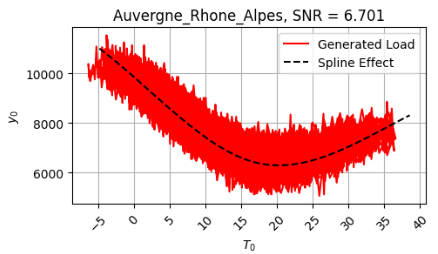}
        \caption{Load generated with a cubic spline basis of rank 10 in Auvergne-Rhône-Alpes.}
    \end{subfigure}
    \caption{Generated temperature and load using pairwise influence between the regions ($\mathbf{\Sigma} = \boldsymbol{\rho}(\mathbf W_\lambda)$).}
    \label{fig: gendata}
\end{figure}


\smallskip
\label{subsec: real-dataset}
\inlinetitle{Real Dataset}{.}~%
We also tested the GNNs on regional data provided by RTE (Réseau de Transport d'Electricité), which is the electricity transmission system operator of France. We enriched the dataset with a variety of features related to electricity consumption and weather conditions in France. It includes generic calendar columns such as an identifier index (used as a trend), multiple date formats, month and year indicators, time of day and year, week number, type of day (weekdays, bank holidays, weekends), specific holiday periods like national, Christmas, summer holidays and a binary indicator for daylight saving time changes. The targeted feature in this dataset is the French national load (in megawatts). Table \ref{tab: vars} gives the description of the features in the dataset.

\begin{table}[h!]
    \centering
    \caption{Features in the dataset.}
    \scalebox{.9}{
    \begin{tabular}{ |c|c| }
        \hline
        Variable & Definition \\ \hline\hline
        \texttt{Date} & Date \\
        \texttt{Region}  & Region  \\
        \texttt{Load} & Electricity consumption (in MW) \\ 
        \texttt{Nebulosity} & Cloud cover \\
        \texttt{Wind} & Wind \\
        \texttt{Temperature} & Temperature (in °C) \\
        \texttt{TempMin}, \texttt{TempMax} & Minimal and maximal values of \texttt{Temp} for the day \\
        \texttt{TempSmoothHigh}/\texttt{Low} & Exponentially smoothed temperatures \\
        \texttt{Instant} & Instant in the day \\ 
        \texttt{Posan} & Position of the day in the year \\ 
        \texttt{DayType} & Categorical variable indicating the type of the day \\
        \texttt{Weekend} & Categorical variable for the weekend \\
        \texttt{Summer}, \texttt{Christmas} & Categorical variable for summer and Christmas \\
        \texttt{Holiday\_zone} & Categorical variable for the other holidays\\
        \hline
        \end{tabular}}
    \label{tab: vars}
\end{table}

\subsection{Experimental Settings}
We place ourselves in a regression setting: we want to forecast the electricity consumption at each time instant $t$ based on features measured at that time.

\smallskip
\inlinetitle{Training Procedure}{.}~%
Denote by $\bX \in \RR^{n\times d\times T}$ and $\by \in \RR^{n\times T}$, respectively, the feature and the target vectors, where $n \in \NN$ is the number of nodes in the graph, $d \in \NN$ is the number of features, and $T \in \NN$ is the number of timesteps to predict. We scale the vectors such that the range of the values is in $[0,1]$ and the scaled versions of the input vector is denoted by $\tilde\bX$. We want to find the best set of parameters $\theta^*$ that minimizes the global loss $\ell$ such that $\theta^* \in {\arg\min_\theta}~\ell(\Phi_\theta(\G,\tilde{\bX}),~ \by)$, if such a solution exists. In practice, we seek to optimize the mean squared error loss defined by:
\begin{equation}\label{eq:MSE}
    \mathsf{RMSE}(\by, \hat\by) = \sqrt{\dfrac{1}{T}\sum_{t=1}^T\Big(\sum_{i=1}^n \by_{i,t} - \hat\by_{i,t}\Big)^2}.
\end{equation}
The set of parameters is computed iteratively using stochastic optimization with batch sampling, and in particular we used the Adam algorithm \cite{kingma2014adam}. 

\smallskip
\inlinetitle{Evaluation Metrics}{.}~%
We used traditional loss functions: Mean Absolute Percentage Error
(MAPE) and Root Mean Square Error (RMSE), which are commonly used to evaluate the performance of forecasting models. MAPE measures the average percentage difference between the forecasted and actual load values, providing an indication of the relative error:
\begin{equation}
    \mathsf{MAPE}(\by, \hat \by) = \dfrac{1}{T}\sum_{t=1}^T \left|{\textstyle\dfrac{\sum_{i=1}^n \by_{i,t} - \hat\by_{i,t}}{\sum_{i=1}^n \by_{i,t}}}\right|.
\end{equation}
RMSE measures the square root of the mean squared difference between the forecasted and actual load values, i.e. the square root of Equation \ref{eq:MSE}. MAPE is chosen for its interpretability as it provides a percentage error that is easy to understand and compare across different scales. This is particularly useful in electricity forecasting as stakeholders need clear insights into the accuracy of predictions relative to actual demand. RMSE, on the other hand, is sensitive to large errors, making it a suitable metric for applications where minimizing significant deviations is important \cite{hyndman2006another}.

\smallskip
\inlinetitle{Parametrization}{.}~%
As with conventional neural networks, learning the parameters and hyper-parameters is a crucial step in the algorithm's performance. Several graph structures were explored: in the real case, dimension reduction and graph fusion methods were tested coupled with the GL3SR and DTW algoritms; in the synthetic case, as the only available feature was the generated temperature, neither fusion nor dimension reduction was necessary. Hyperparameters were optimized using a simple grid search over the space $(\texttt{batch\_size},~\texttt{n\_layers},~ \texttt{hidden\_channels},~\texttt{n\_epochs}) \in \{256, 512, 1024\}\times\{3,4,5,12\} \times\{10, 32, 50, 64, 128\}\times\{300\}$. We considered two graph-based models: a GCN and a SAGE. The results of the grid searche are shown in Tables \ref{table: gs1}, \ref{table: gs2}, and \ref{table: gs3}.

\begin{figure}[h!]
    \centering
    \begin{subfigure}[b]{.49\textwidth}
        \centering
        \includegraphics[width=.98\linewidth]{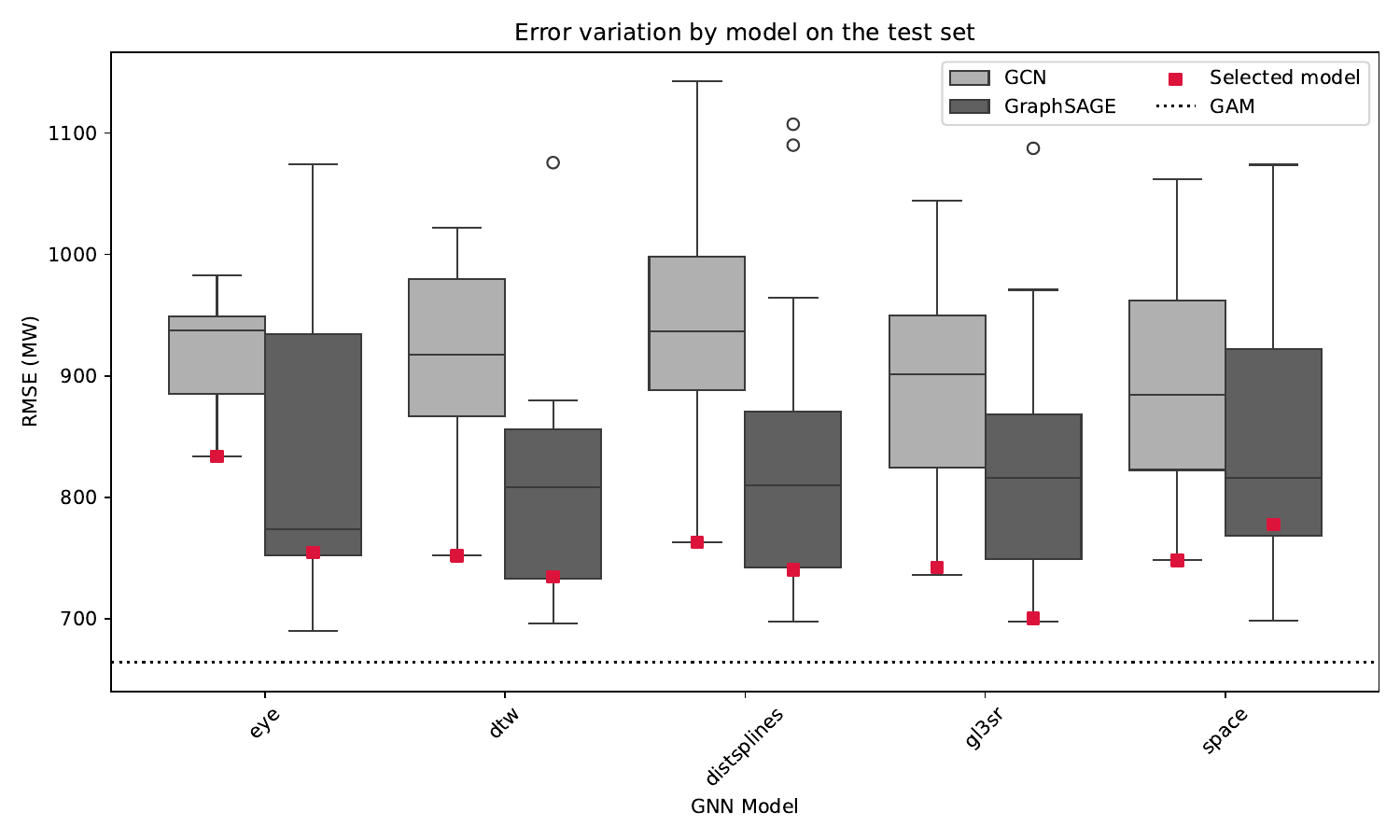}
        \caption{$\mathbf{\Sigma} = \boldsymbol{\rho}(\mathbf W_\lambda)$.}
        \label{fig: boxplot}
    \end{subfigure}\hfill
    \begin{subfigure}[b]{.49\textwidth}
        \centering
        \includegraphics[width=.98\linewidth]{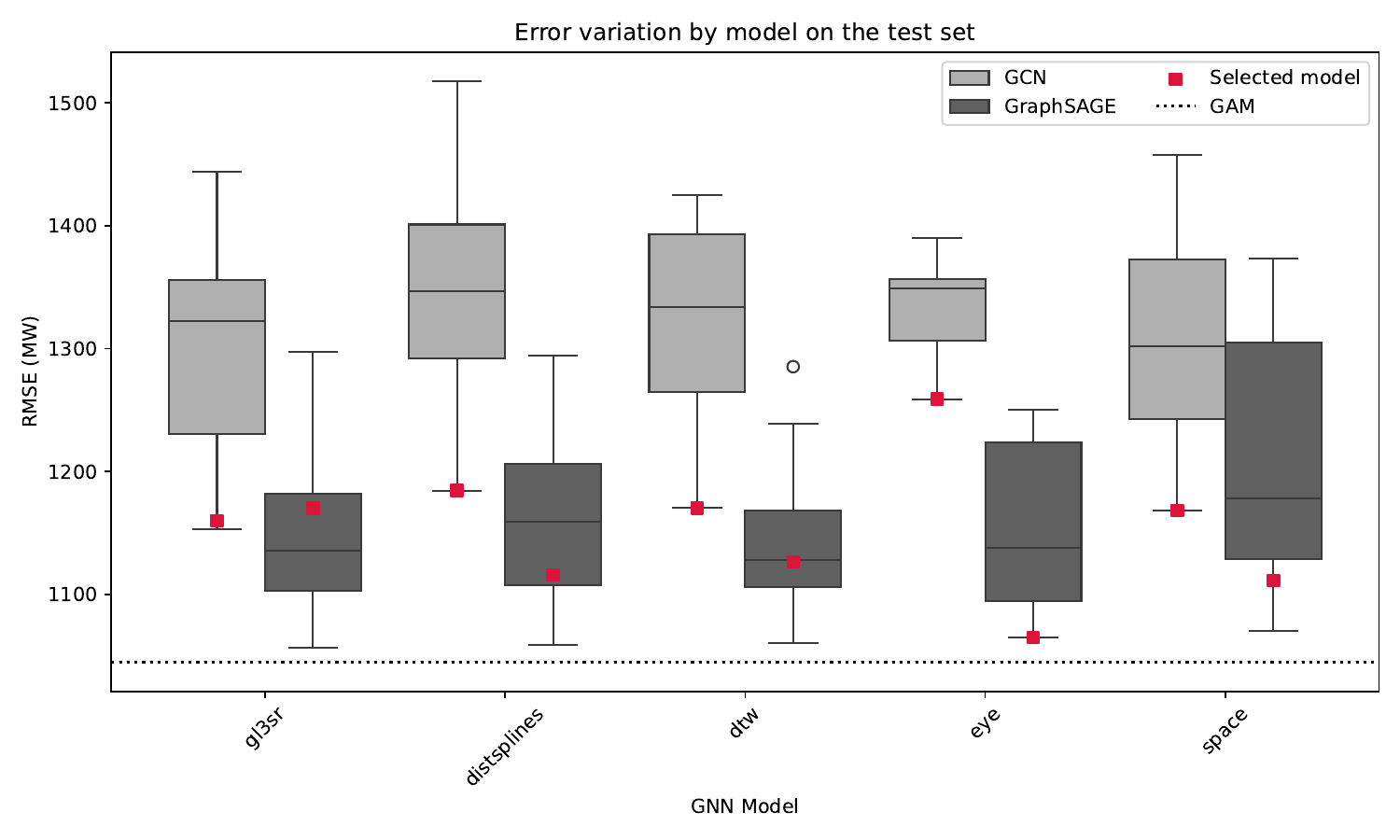}
        \caption{$\mathbf{\Sigma} = \boldsymbol{I}$.}
        \label{fig: boxplot_eye}
    \end{subfigure}
    \centering
    \caption{Error variation by model on the synthetic test set.}
\end{figure}

Note that the hyperparameters are similar from one simulation to the other, but differ greatly on the real dataset. Also, the SAGE model converges faster on average than the GCN, hence the difference in average training time. However, for both GCN and SAGE, the inference time is less than a second. Once the optimization has been carried out for all GNN/weight matrix pairs, the models are selected by taking the sets of hyperparameters that minimize the loss on the validation set, allowing good generalization to the models. On Figure \ref{fig: boxplot}, the distribution of errors per model on the test set with the correlation structure is represented, as well as the performance of the selected model (which is not necessarily the model with the lowest error on this set). This figure shows that the selected models are in most cases in the top 25\% of the best models on the test set. We can also note a difference between GCNs and SAGEs: the best GCNs models on the validation and test sets are in the majority confounded, whereas SAGEs underperform on the test set, while still outperforming GCNs on average.

\begin{table}[h!]
    \caption{Hyperparameters of the GNN models for the real dataset.}
    \label{table: gs1}
    \centering\footnotesize
    \scalebox{.9}{
    \begin{tabular}{c|c|c|c|c|c}
        \hline
        Model & Graph structure & \texttt{batch\_size} & \texttt{n\_layers} & \texttt{hidden\_channels} &
        \texttt{n\_epochs}\\
        \hline\hline
        GCN & Identity & 512 & 12 & 10 & 6 \\
        GCN & Space & 256 & 3 & 64 & 293 \\
        GCN & DistSplines & 256 & 4 & 64 & 213 \\
        GCN & GL3SR & 512 & 12 & 10 & 6 \\
        GCN & DTW & 512 & 12 & 32 & 191 \\
        \hline\hline
        \rowcolor{gray!30}SAGE & Identity & 512 & 12 & 128 & 2\\
        \rowcolor{gray!30}SAGE & Space & 256 & 4 & 64 & 211 \\
        \rowcolor{gray!30}SAGE & DistSplines & 1024 & 3 & 128 & 249 \\
        \rowcolor{gray!30}SAGE & GL3SR & 256 & 12 & 128 & 2 \\
        \rowcolor{gray!30}SAGE & DTW & 1024 & 12 & 50 & 20\\
        \hline
    \end{tabular}}
\end{table}

\begin{table}[h!]
    \caption{Hyperparameters of the GNN models for the synthetic dataset ($\mathbf{\Sigma} = \boldsymbol{\rho}(\mathbf W_\lambda))$.}
    \label{table: gs2}
    \centering\footnotesize
    \scalebox{.9}{
    \begin{tabular}{c|c|c|c|c|c|c|c}
        \hline
        Model & Graph structure & \texttt{batch\_size} & \texttt{n\_layers} & \texttt{hidden\_channels} &
        \texttt{n\_epochs} &
        \# parameters &
        Training time\\
        \hline\hline
        GCN & Identity & 512 & 3 & 50 & 127 & 2701 & $\sim 390s$\\
        GCN & Space & 1024 & 3 & 64 & 158 & 4353 & $\sim 480s$ \\
        GCN & DistSplines & 1024 & 3 & 64 & 168 & 4353 & $\sim 510s$ \\
        GCN & GL3SR & 1024 & 3 & 64 & 147 & 4353 & $\sim 450s$\\
        GCN & DTW & 1024 & 3 & 64 & 146 & 4353 & $\sim 450s$\\
        \hline\hline
        \rowcolor{gray!30}SAGE & Identity & 512 & 4 & 50 & 9 & 10351 & $\sim 50s$ \\
        \rowcolor{gray!30}SAGE & Space & 512 & 4 & 50 & 12 & 10351 & $\sim 60s$\\
        \rowcolor{gray!30}SAGE & DistSplines & 512 & 4 & 50 & 11 & 10351 & $\sim 50s$ \\
        \rowcolor{gray!30}SAGE & GL3SR & 256 & 3 & 50 & 3 & 5301 & $\sim 10s$ \\
        \rowcolor{gray!30}SAGE & DTW & 512 & 4 & 50 & 12 & 10351 & $\sim 50s$ \\
        \hline
    \end{tabular}}
\end{table}

\begin{table}[h!]
    \caption{Hyperparameters of the GNN models for the synthetic dataset ($\mathbf{\Sigma} = \boldsymbol{I})$.}
    \label{table: gs3}
    \centering\footnotesize
    \scalebox{.9}{
    \begin{tabular}{c|c|c|c|c|c|c|c}
        \hline
        Model & Graph structure & \texttt{batch\_size} & \texttt{n\_layers} & \texttt{hidden\_channels} &
        \texttt{n\_epochs} &
        \# parameters &
        Training time \\
        \hline\hline
        GCN & Identity & 512 & 3 & 50 & 127 & 2701 & $\sim 390s$ \\
        GCN & Space & 1024 & 3 & 64 & 158 & 4353 &$\sim 480s $\\
        GCN & DistSplines & 1024 & 3 & 64 & 168 & 4353 &$\sim 510s$ \\
        GCN & GL3SR & 1024 & 3 & 64 & 147 & 4353 &$\sim 450s$ \\
        GCN & DTW & 1024 & 3 & 64 & 146 & 4353 &$\sim 450s$ \\
        \hline\hline
        \rowcolor{gray!30}SAGE & Identity & 512 & 4 & 50 & 9 & 10351 &$\sim 50s$ \\
        \rowcolor{gray!30}SAGE & Space & 1024 & 3 & 64 & 21 & 8577 &$\sim 60s$  \\
        \rowcolor{gray!30}SAGE & DistSplines & 1024 & 3 & 50 & 6 & 5301 &$\sim 15s$ \\
        \rowcolor{gray!30}SAGE & GL3SR & 1024 & 4 & 50 & 10 & 10351 &$\sim 40s$\\
        \rowcolor{gray!30}SAGE & DTW & 512 & 4 & 50 & 13 & 10351 &$\sim 50s$  \\
        \hline
    \end{tabular}}
\end{table}

On Figure \ref{fig: boxplot_eye}, the variation of errors by model on the uncorrelated test set is shown. This time, we observe that the selected SAGE models are less often in the top 25\% (unlike the GCNs), while still performing better than the GCNs, and the gap between the two model types is more pronounced than for the first dataset.

\subsection{Results}


\smallskip
\inlinetitle{Numerical Results}{.}~%
The national-level numerical results for France are presented in Table \ref{table: national-results}. These results include both the national dataset and synthetically generated datasets. Generally, across the two tested models, the SAGE model performs better than GCN on average, which is a finding that is consistent with the related literature on the subject. Additionally, it is evident that for a given GNN model, the results vary depending on the input graph structure: on the real dataset, considering a purely autoregressive structure (corresponding to the identity matrix), the results are significantly worse than when adding relationships between regions. This validates the hypothesis that there is useful information in regional links and highlights the importance of applying graph-based modeling to the electricity forecasting problem. The results on synthetic datasets also show that, when the regions are influenced by one another, the gain of GNNs compared to an optimal model (here the GAM) is greater than that obtained on data with no correlations (it should be noted that the variance is greater when the covariance matrix is the identity matrix, so poorer performance is expected).

\begin{figure}[H]
    \centering
    \includegraphics[width=\textwidth]{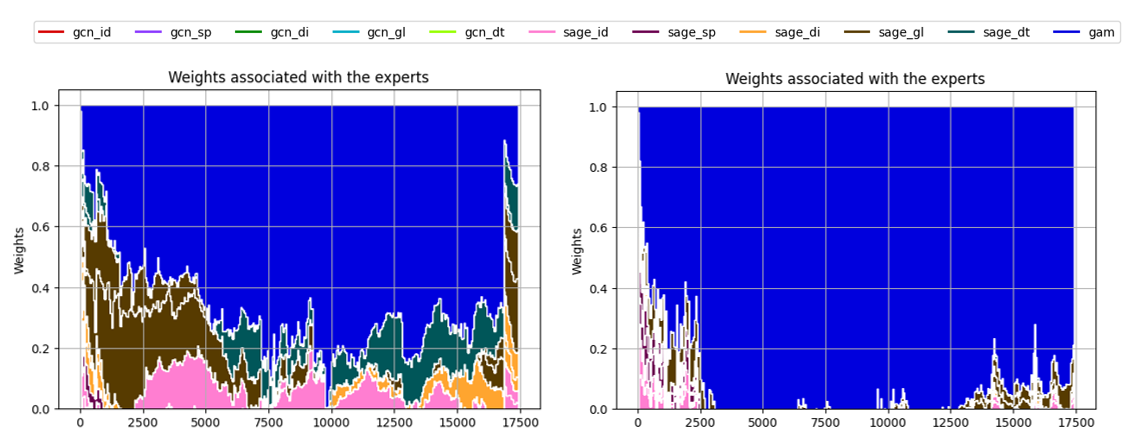}
    \caption{Weights associated with the experts on the synthetic datasets. $\mathbf{\Sigma} = \boldsymbol{\rho}(\mathbf W_\lambda)$ (left), $\mathbf{\Sigma} = \boldsymbol{I}$ (right). GAM is the main expert, followed by mutliple SAGE-\texttt{gl3sr} and SAGE-\texttt{dtw}.}
    \label{fig: aggreg_res}
\end{figure}

\begin{table}[t!]
    \caption{Numerical performance in MAPE (\%) and RMSE (MW) at national level on the test set.}
    \label{table: national-results}
    \centering\footnotesize
    \scalebox{.8}{
    \begin{tabular}{c|c|c|c}
        \hline
        Model & \specialcell{\multicolumn{1}{c}{Real Dataset} \\ 
            MAPE (\%) \quad RMSE (MW) } & \specialcell{\multicolumn{1}{c}{Synthetic Dataset ($\mathbf{\Sigma} = \boldsymbol{\rho}(\mathbf W_\lambda))$} \\ MAPE (\%) \quad RMSE (MW) } & \specialcell{\multicolumn{1}{c}{Synthetic Dataset ($\mathbf{\Sigma} = \boldsymbol{I}$)} \\ MAPE (\%) \quad RMSE (MW) } \\
        \hline\hline
        \rowcolor{orangecb!40}GAM-Regions & \textbf{1.48} \quad\quad \textbf{1018} & \textbf{1.11} \quad\quad \textbf{662} & \textbf{1.75} \quad\quad \textbf{1043}  \\
        \rowcolor{orangecb!40}Feed Forward & 1.54 \quad\quad 1071 & 3.82 \quad\quad 3141 & 4.49 \quad\quad 3213  \\
        \hline\hline
        \rowcolor{bluecb!40}GCN-\texttt{identity} & 5.66 \quad\quad 3949 & 1.43 \quad\quad 834 & 2.16 \quad\quad 1259 \\
        \rowcolor{bluecb!40}GCN-\texttt{space} & 2.07 \quad\quad 1452 & 1.26 \quad\quad 749 & 1.98 \quad\quad 1169 \\
        \rowcolor{bluecb!40}GCN-\texttt{distsplines} & 2.04 \quad\quad 1404 & 1.29 \quad\quad 764 & 2.01 \quad\quad 1185 \\
        \rowcolor{bluecb!40}GCN-\texttt{gl3sr} & 5.95 \quad\quad 4210 & \textbf{1.25} \quad\quad \textbf{743} & \textbf{1.97} \quad\quad \textbf{1160} \\
        \rowcolor{bluecb!40}GCN-\texttt{dtw} & \textbf{1.82} \quad\quad \textbf{1276} & 1.26 \quad\quad 753 & 1.99 \quad\quad 1171 \\
        \hline 
         \rowcolor{bluecb!40}SAGE-\texttt{identity} & 4.38 \quad\quad 3021 & 1.25 \quad\quad 755 & \textbf{1.78} \quad\quad \textbf{1066} \\
        \rowcolor{bluecb!40}SAGE-\texttt{space} & 1.96 \quad\quad 1350 & 1.29 \quad\quad 778 & 1.85 \quad\quad 1112 \\
        \rowcolor{bluecb!40}SAGE-\texttt{distsplines} & 2.06 \quad\quad 1410 & 1.22 \quad\quad 741 & 1.84 \quad\quad 1116 \\
        \rowcolor{bluecb!40}SAGE-\texttt{gl3sr} & \textbf{1.78} \quad\quad \textbf{1234} & \textbf{1.15} \quad\quad \textbf{701} & 1.92 \quad\quad 1171 \\
        \rowcolor{bluecb!40}SAGE-\texttt{dtw} & 1.90 \quad\quad 1335 & 1.21 \quad\quad 735 & 1.86 \quad\quad 1127 \\
        \hline\hline
        \rowcolor{orangecb!40}Mixture (Baseline) & 1.31 \quad\quad 925 & 1.11 \quad\quad 662 & 1.76 \quad\quad 1044  \\
        \rowcolor{bluecb!40}Mixture (GNNs) & 1.48 \quad\quad 1092 & 1.12 \quad\quad 677 & 1.98 \quad\quad 1171 \\
        \rowcolor{darkred!40}Mixture (Baseline + GNNs) & \textbf{1.13} \quad\quad \textbf{844} & \textbf{1.08} \quad\quad \textbf{647} & \textbf{1.76} \quad\quad \textbf{1050} \\
        \hline
    \end{tabular}}
\end{table}
The aggregation of experts in Figure \ref{fig: aggreg_res} reveals two key-insights: first, GNNs introduce diversity when there is an underlying graph structure in the data; second, the weight of GNNs decreases in the later part of the year, suggesting that separate models should be trained for summer and winter. The analysis of the results on real data shows that although GAMs perform better than GNNs, the latter seem significant in the mixture of experts, since they bring diversity and improve the RMSE by around 200MW, which is consistent with the analysis of the results on the synthetic data. Important to also note that the results achieved using the correlation matrix of distances between splines are less effective on the real dataset. This outcome is somewhat expected, because while the temperature is a significant variable, the actual consumption depends on additional weather and calendar factors that were not considered in the construction of the splines. The matrices that give the best results on average are those obtained after the DTW and GL3SR algorithms, both on the real dataset and on the dataset with correlations between regions, for both GCN and SAGE. On the uncorrelated dataset, on the other hand, we find that the identity matrix gives the best results, which is again in line with the shape of the data.

\smallskip
\inlinetitle{Explainability Results}{.}~%
We constructed explanatory subgraphs at several times of the year (January, June, August, and November), each using data during one day. To do this, we summed and normalized all explanatory subgraphs at half-hourly step for a given day. Two datasets were considered: the real dataset and the dataset with the correlation structure ($\mathbf{\Sigma} = \boldsymbol{\rho}(\mathbf W_\lambda))$. \textsf{GNNExplainer} was trained to learn a mask $\mathbf{M}$, allowing the sub-graph to be represented as $\mathbf{W_\lambda} \odot \sigma(\mathbf M)$, where $\odot$ denotes the Hadamard product and $\sigma$ is a sigmoid function that maps $\mathbf M$ to $[0,1]^{n\times n}$.

\begin{figure}[t!]
    \centering
    \begin{subfigure}[b]{.49\textwidth}
        \centering
        \includegraphics[height=160px]{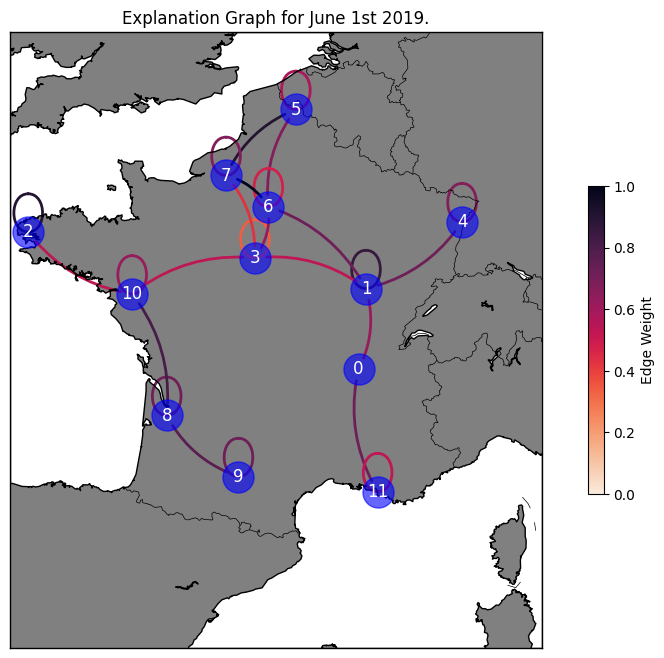}
    \caption{Synthetic dataset ($\mathbf{\Sigma} = \boldsymbol{\rho}(\mathbf W_\lambda)$).}
    \end{subfigure}\hfill
    \begin{subfigure}[b]{.49\textwidth}
        \centering
        \includegraphics[height=160px]{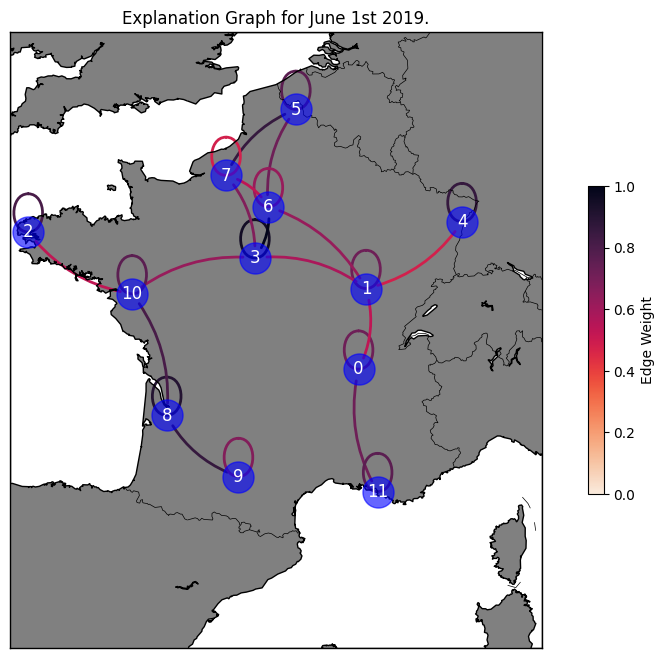}
        \caption{Real dataset.}
    \end{subfigure}
    \caption{Explanation graphs in June 2019 obtained from the \texttt{space} matrix.}
    \label{fig: exp_graphs}
\end{figure}

In Figure \ref{fig: exp_graphs}, only the explanatory subgraphs for June are displayed, since the results were similar across the different months of the year. These subgraphs, when compared with the graph in Figure \ref{fig: exponential-weight-matrix}, reveal that the autoregressive component in updating region representations is less significant than initially assumed. This underscores the advantage of aggregating neighborhood representations, notably, there are strong connections between regions along the Atlantic seaboard, excluding Brittany. Other significant connections emerge, such as links between geographically close regions like Hauts-de-France, Normandy, and Ile-de-France. Additionally, the behavior of the Centre-Val-de-Loire region appears to diverge between simulation and actual data: in one case, the autoregressive aspect is weak, while in the other, it is the opposite. This can be explained partly by the increased difficulty in forecasting consumption in this region, as it has the lowest consumption levels in the territory.

\section{Discussion}

The experiments are set in a regression context for a given instant: future experiments could aim to integrate previous instants into the forecast for the target instant, e.g. by exploiting temporal GNN models \cite{rossi2020temporal}.
The inference methods presented here could also be coupled with explainability methods, and could therefore be split into two steps: first, ``classical'' inference as presented in this work, then multiplication of the inferred matrix by a mask. Future experiments could also include renewable production data (solar and wind) to take into account the evolution of the grid towards a decentralized network.
Considering the results obtained with distance-based graphs (such as DTW and spline effect distances), we plan to derive other graphs based on socio-economic data. Additionally, we intend to use normalized load data to prevent regions with similar level of average consumption from being considered as related.

\begin{figure}[t!]
    \centering
    \includegraphics[width=\textwidth]{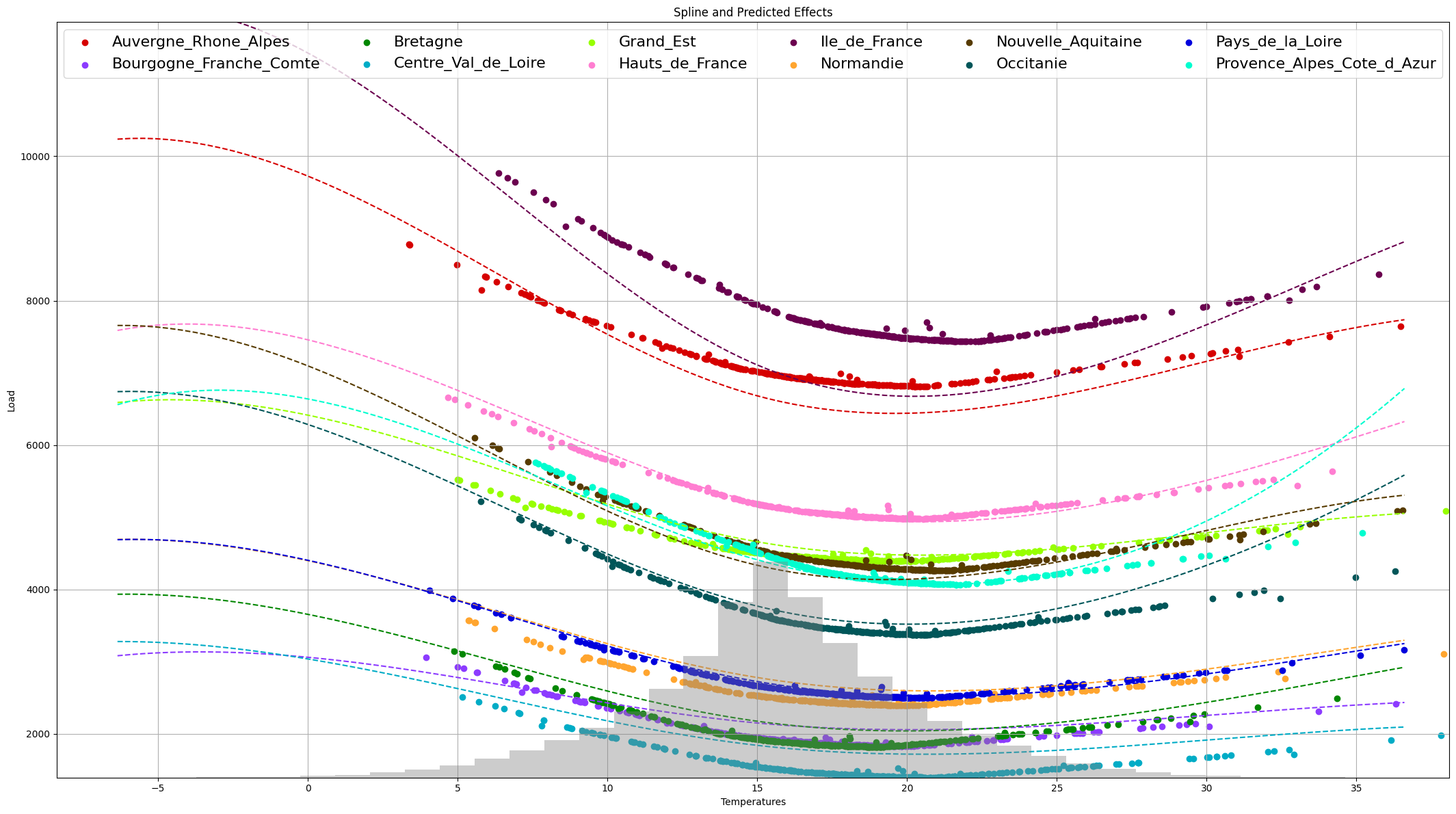}
    \caption{Spline (dashed line) and predicted (scatter plot) effects. The distribution of the generated temperatures is represented in gray.}
    \label{fig: aleplot}
\end{figure}

Figure \ref{fig: aleplot} presents the Accumulated Local Effects (ALE) plots \cite{apley2020visualizing} illustrating the impact of temperature on load for each region. The reconstructed effects generally align well with the effects generated by the splines, although the air-conditioning and heating effects at extreme temperatures are not accurately captured. This discrepancy can be attributed to the Gaussian distribution of the synthetic 
temperatures, where extremes are located at the distribution tails. A potential improvement could be to give more weight to situations with extreme temperatures, or to learn patterns specifically for summer and winter, which is a conclusion that is in line with the results in Figure~\ref{fig: aggreg_res}.

In order to apply distance algorithms on univariate signals for creating graphs, we used SVD. This approach can be limiting in our application, since we retain only the eigenvalue of higher magnitude, which introduces a bias towards the associated feature. Another way of doing this is to construct as many graphs as there are features and then merge them, for example by choosing a graph from the convex space generated by the $d$ graphs, which will be done in a future  work.

Finally, to consolidate the results on the explainability of GNNs, we plan to develop Graph Attention Networks, both on a spatial level as done in \cite{feik2024graph}, and on a temporal level.

\section{Conclusion}

In this paper, we introduced a method for constructing graphs tailored to the problem of electricity consumption forecasting. Through diverse inference methods, graph fusion, and the aggregation of different models, the final graph is robust and enhances the performance of GAMs when the data has an underlying graph structure. The explanatory sub-graphs help understand the connections between different regions, validate or refute hypotheses about these links, and consequently propose a corrected version of the input graph. Areas for improvement include incorporating a temporal structure into the models and developing specialized models for different times of the year, such as winter and summer. Ultimately, the outcomes achieved with GNNs show promise: in cases where data exhibits a graph-based structure, these models perform much better, contributing to diverse enhancements through improved expert aggregation.

\smallskip
\begin{credits}
\inlinetitle{Disclosure of Interests}{.}~%
The authors have no competing interests to declare that are relevant to the content of this article.
\end{credits}
%
%
%
%
\bibliography{main}
\bibliographystyle{splncs04}
\end{document}